\documentclass{article}

\PassOptionsToPackage{numbers, compress}{natbib}
\usepackage[final]{neurips_2024}
\usepackage[utf8]{inputenc}
\usepackage[T1]{fontenc}
\usepackage{hyperref}
\usepackage{url}
\usepackage{booktabs}
\usepackage{amsfonts}
\usepackage{nicefrac}
\usepackage{microtype}
\usepackage{xcolor}
\usepackage{makecell}
\usepackage{graphicx}

\title{BarcodeMamba: State Space Models\\ for Biodiversity Analysis}
\author{
  Tiancheng Gao\textsuperscript{1,2}, Graham W.~Taylor\textsuperscript{1,2*} \\
  \textsuperscript{1}University of Guelph \\
  \textsuperscript{2}Vector Institute for AI \\
}

\begin{document}
\maketitle
\def\thefootnote{$*$}\footnotetext{Author for correspondence: \texttt{gwtaylor@uoguelph.ca}}

\begin{abstract}
DNA barcodes are crucial in biodiversity analysis for building automatic identification systems that recognize known species and discover unseen species. Unlike human genome modeling, barcode-based invertebrate identification poses challenges in the vast diversity of species and taxonomic complexity. Among Transformer-based foundation models, BarcodeBERT excelled in species-level identification of invertebrates, highlighting the effectiveness of self-supervised pretraining on barcode-specific datasets. Recently, structured state space models (SSMs) have emerged, with a time complexity that scales sub-quadratically with the context length. SSMs provide an efficient parameterization of sequence modeling relative to attention-based architectures. Given the success of Mamba and Mamba-2 in natural language, we designed BarcodeMamba, a performant and efficient foundation model for DNA barcodes in biodiversity analysis. We conducted a comprehensive ablation study on the impacts of self-supervised training and tokenization methods, and compared both versions of Mamba layers in terms of expressiveness and their capacity to identify ``unseen'' species held back from training. Our study shows that BarcodeMamba has better performance than BarcodeBERT even when using only 8.3\% as many parameters, and improves accuracy to 99.2\% on species-level accuracy in linear probing without fine-tuning for ``seen'' species. In our scaling study, BarcodeMamba with 63.6\% of BarcodeBERT's parameters achieved 70.2\% genus-level accuracy in 1-nearest neighbor (1-NN) probing for unseen species. The code repository to reproduce our experiments is available at \url{https://github.com/bioscan-ml/BarcodeMamba}.

\end{abstract}

\section{Introduction}

A DNA barcode is a short and standardized section of nucleotides within the genome that allows taxonomic identification at the species level without the need to consider entire genomes, making it efficient and invaluable for biodiversity analysis \cite{barcode}. For many animal groups, particularly invertebrates, part of the mitochondrial gene Cytochrome $c$ oxidase Subunit I (COI) \cite{coigene} is commonly used. However, different genes serve as barcodes for other organisms. Plants often rely on plastid genes such as rbcL and matK, while for fungi, the internal transcribed spacer (ITS) region is frequently employed. These genetic markers can be utilized to establish automatic taxonomic identification systems that recognize species known and unknown to science. Such systems significantly reduce the amount of manual labor typically required by taxonomic experts.

Among the barcode-based analysis tasks, invertebrate taxonomic classification \cite{canadianinvertebrates, bolddataset} is particularly challenging due to the imbalance in data distributions and intrinsic diversity of labels. Identifying taxonomic relationships among a large number of classes is complex and requires expertise in taxonomy. Unidentified species and incomplete taxonomic annotations pose challenges for accurate classification. Therefore, this task differs significantly from the design objectives of most modern DNA models.

Numerous studies have been proposed to tackle the challenges posed by DNA analysis and genomics. Early machine learning approaches employed task-specific end-to-end training based on convolutional neural networks (CNNs) \cite{cnnencoder}. These methods yield models capable of solving classification tasks with high accuracy using a relatively small number of parameters.
In recent years, Transformers \cite{transformer} have dominated various sequence modeling tasks, notably in natural language. Their ability to leverage self-supervised learning (SSL) on unlabeled datasets and fine-tune on downstream tasks has made them highly effective. Transformer-based foundation models have been introduced into the genomics space \cite{transformersandbeyond,bendbenchmark}, bringing their ability to generalize across diverse tasks. 
Models like DNABERT \cite{dnabert} and DNABERT-2 \cite{dnabert2} have demonstrated this capability in human and multi-species DNA analysis, as well as the Nucleotide Transformer \cite{nucleotidetransformer}, GENA-LM \cite{GENA_LM} and GROVER \cite{grover}.
However, these models were not specifically designed to address the challenges posed by biodiversity analysis. While BERTax \cite{bertax} can be fine-tuned for taxonomic classification, its predictions are limited to known taxa and only at the superkingdom, phylum and genus levels.

To fill this gap, BarcodeBERT \cite{barcodebert} was developed as a specialized model for DNA barcode analysis, with a particular focus on challenges posed by species classification of invertebrates. Unlike its predecessors, BarcodeBERT was designed to account for the unique characteristics of DNA barcodes. In particular, the use of non-overlapping $k$-mer-based tokenizers demonstrated significant improvements in zero-shot classification of unseen species to the correct genus, surpassing the performance of CNNs and off-the-shelf Transformer-based DNA foundation models.
Recently, foundation models utilizing a structured SSM as their backbone have demonstrated impressive performance in human DNA modeling \cite{hyenadna,mambadna}. Nevertheless, consistent with BarcodeBERT's results, we find that current off-the-shelf foundation models may not perform optimally without barcode-specific pretraining.

In this study, we introduce BarcodeMamba, an efficient foundation model for DNA barcode modeling. Our model demonstrates competitive performance compared to BarcodeBERT on the Canadian Invertebrate species classification task with only 8.3\% of the parameters. BarcodeMamba reaches 99.2\% accuracy on a species-level linear probing task without fine-tuning, demonstrating its capability in DNA barcode modeling. After scaling up, BarcodeMamba achieves 99.2\% accuracy on species-level linear probing and 70.2\% on 1-NN genus-level probing.

The main contributions of this paper are:
\begin{enumerate}
    \item Introducing BarcodeMamba, an efficient method for self-supervised learning using DNA barcode data for biodiversity analysis based on the state-of-the-art Mamba-2 architecture.
    \item Conducting a comprehensive ablation study to identify the optimal settings for different aspects of biodiversity analysis, including character-level and $k$-mer tokenizers and various tasks for self-supervised pretraining. Comparing both versions of Mamba \cite{mamba,mamba2} to determine their respective advantages in modeling DNA barcodes.
    \item Scaling the top two BarcodeMamba variants to assess improvements in both DNA barcode modeling (measured by perplexity) and downstream classification tasks (species- and genus-level accuracy) under both tokenization strategies.
    \item Comparing BarcodeMamba's performance with baselines from classical supervised learning, as well as Transformer-based and SSM-based foundation models, in the taxonomic classification of 1.5\,M Canadian invertebrates.
\end{enumerate}
\vspace{-0.1cm}
\section{Background: Structured State Space Models for DNA Analysis}
To address the quadratic cost of self-attention in Transformer-based models and the need to handle long contexts, SSMs have been developed to build sequential models with linear or near-linear complexity. This advancement has significantly reduced computation costs and accelerated training speed. Since the emergence of the structured state space sequence (S4) model \cite{s4}, SSMs can be computed with a long convolution during training and recurrence during inference, enabling more efficient computations for sequence modeling. Furthermore, these models exhibit promising properties when scaled up similarly to Transformers \cite{scalelaw}. 

Unlike prior Linear Time Invariant (LTI) models, Mamba-based models are capable of effective unidirectional representation learning. Mamba \cite{mamba} has been proposed as a linear-time sequence model. Most recently, Mamba-2 \cite{mamba2} was introduced to integrate the theory of SSMs with attention mechanisms, increasing the efficiency of modeling sequences of dense information, such as language. The selective copying synthetic task introduced in Mamba demonstrates that Mamba-based models can use input-dependent parameterization to selectively remember or ignore inputs based on their content \cite{mamba}.

Building upon this property, we expect Mamba-based models to excel at handling nucleotide alignment gaps in DNA barcode sequences. This makes representation learning less susceptible to variations in DNA sample quality, sequencing technologies, and specific regions of genomes with structural complexity that are difficult to identify due to technical limitations. Furthermore, the results of the multi-query associative recall synthetic task indicate that Mamba-2 is able to memorize multiple associations, and efficiently parameterize and parallelize its implementation for improved performance in modeling dense information \cite{mamba2}.
Additionally, Mamba-based models are capable of achieving competitive results compared to Transformer-based models of the same size or larger in language modeling. Motivated by this, we developed a Mamba-2-based DNA barcode foundation model to explore its potential in biodiversity analysis. With a dual form of kernelized attention and linear recurrence in Mamba-2, BarcodeMamba can be efficiently trained with hardware-aware parallelization and inferred auto-regressively.

\vspace{-0.2cm}
\section{Method}

This section presents an overview of the DNA barcode dataset used in our experiments and describes the architecture of BarcodeMamba, along with the baseline models used for comparison.
\vspace{-0.1cm}
\subsection{Dataset}

In this study, we employed the Canadian invertebrate dataset, consisting of 1.5\,M samples from the Barcode of Life Datasystem (BOLD) as our primary data source \citep{canadianinvertebrates}. We adopted the preprocessing method introduced in BarcodeBERT \cite{barcodebert}. Each record in the dataset consists of five possible characters, namely A, T, G, C, and N, representing alignment gaps or IUPAC ambiguity codes. We examined two tokenizers used in DNA barcode modeling: $k$-mer, used by BarcodeBERT, and character-level, which is popular in (non-barcode) SSM models.

During both self-supervised pretraining and downstream evaluation phases, we applied the same data splits as in BarcodeBERT\cite{barcodebert}. The length of all DNA barcode data was fixed at 660 base pairs of nucleotides. During self-supervised pretraining, 95\% of the data, consisting of 0.9\,M sequences, was used for training and 5\% (47.1\,k sequences) for validation. After pretraining, we fine-tuned the model for species-level classification of known arthropods using a dataset comprising 1,653 classes. During fine-tuning on 67.2\,k sequences, 70\% of the data was used for training, 20\% for testing, and 10\% for validation. In addition to probing unseen species as in BarcodeBERT, we measured the perplexity of the model's output on unseen data that did not overlap with the pretraining or fine-tuning subsets.

\subsection{Network Architectures}
\paragraph{CNN Encoder and Transformer Baselines}

We adopted the experimental setting in BarcodeBERT for our study \cite{barcodebert}. Our CNN and Transformer baselines include a supervised CNN encoder similar to that used in \citep{cnnencoder} and BERT-based foundation models. The CNN encodes the context of DNA data with convolution layers, while DNABERT, designed for genomic understanding, utilized $k$-mer tokenizers to process nucleotide context and effectively predicted splicing and transcription factor binding site in human DNA. In DNABERT-2, the authors deployed Byte Pair Encoding tokenizers for genomic tokenization across multiple species. BarcodeBERT also serves as a baseline in our research, utilizing a $k$-mer tokenizer and implementing direct masked pretraining on barcodes.

\paragraph{State Space Model Baselines}

Our SSM baselines include HyenaDNA and Caduceus models. HyenaDNA used an implicit convolution to match the performance of attention-based transformers in DNA modeling. By leveraging global context at each layer, the authors extended the context length up to 1\,M in human genome modeling \cite{nucleotidetransformer,genomicbenchmark}. In contrast to aggregation for creating vocabularies, a character-level tokenizer was implemented to capture single nucleotide polymorphisms or mutations and dependencies in gene expression. As a decoder-only causal model with a sequence-to-sequence architecture, HyenaDNA utilized next token prediction (NTP) for pretraining. Notably, the model demonstrated superior performance on the benchmarks considered by the nucleotide transformer \cite{nucleotidetransformer} as well as genomic benchmarks \cite{genomicbenchmark}.

Caduceus \cite{mambadna} is a DNA modeling framework that leverages MambaDNA blocks. It utilizes a Bi-Mamba architecture to incorporate bi-directionality for analyzing reverse complementarity (RC) on pairs of DNA strands. Unlike our primary focus on species identification and discovering unseen species, the authors of Caduceus performed efficient variant prediction to study evolutionary pressure. The Mamba computation was applied twice: once on the reversed and once on the forward sequence, with an efficient implementation using shared projection weights. Additionally, masked language modeling (MLM) was used for pretraining. Similar to HyenaDNA, Caduceus tokenizes sequences by characters. The Caduceus-PS setting incorporates RC-equivariant token embedding, while the Caduceus-PH setting involves RC data augmentation. Caduceus outperforms uni-directional models lacking RC equivariance.
\vspace{-0.3cm}
\paragraph{BarcodeMamba}
BarcodeMamba follows a language model backbone and decoder architecture. The model processes input through $n$ stacked blocks, each containing layer normalization, a multi-layer perceptron, and a Mamba-2 mixing layer that maps $d$-dimensional inputs through a $p$-dimensional head. The resulting hidden states serve as input to the decoder. While previous SSM-based foundation models for DNA analysis have primarily relied on character-level tokenizers for human DNA sequences, BarcodeMamba explores both character-level and $k$-mer tokenization approaches. The $k$-mer approach enables the model to capture local patterns essential for classification, rather than processing individual nucleotides. During pretraining, we augmented the data using reverse complement sequences and investigated two pretraining objectives: NTP, which is preferred by causal models, and MLM, which was successfully applied in BarcodeBERT and Caduceus for discriminative downstream tasks.
\vspace{-0.3cm}
\section{Experiments}
To evaluate the performance of a Mamba-2-based architecture in DNA barcode-based biodiversity analysis, we reported various evaluation metrics \cite{sslevaluation} and gradually scaled up \cite{scalelaw} variants of BarcodeMamba to identify further performance improvements. We also compared BarcodeMamba with both supervised and self-supervised baselines from Transformers and SSMs.
\vspace{-0.2cm}
\subsection{Task Definition and Methodology}
Species classification of invertebrates using DNA barcode analysis presents unique challenges given the intricate taxonomic relationships and a vast number of classes. Furthermore, existing datasets are highly imbalanced, and there remain many undiscovered species. Our focus is on DNA barcode-based taxonomic classification, as investigated by BarcodeBERT \cite{barcodebert}. 

Our methodology consists of several key steps:
\begin{enumerate}
    \item {\bf Fine-tuned:} We first train BarcodeMamba on a pretraining dataset split, followed by fine-tuning using supervised training datasets. We then evaluate the models' accuracies on species-level barcode-based classification.
    \item {\bf Linear probe:} To assess the effectiveness of self-supervised learning on DNA barcodes, we employ pretrained models as feature extractors. This involves training a linear classifier on embeddings extracted from each pretrained model, and evaluating its accuracy of classifying known species.
    \item {\bf 1-NN probe:} Finally, to evaluate the model's ability to generalize to new taxonomic groups, we implement genus-level 1-NN probing on barcode sequences from unseen species. This involves training a 1-NN classifier on the embeddings of pretrained models and evaluating its accuracy of identifying unknown species at genus level.
\end{enumerate}

\subsection{Experimental Results}
\subsubsection{Comparison with Baselines}
\paragraph{Implementation Details}
We evaluate BarcodeMamba against a comprehensive set of baselines used in the BarcodeBERT study and recent work on SSM-based DNA foundation models. Our baselines include a traditional CNN encoder \cite{cnnencoder} that is trained by supervised learning, as well as pretrained foundation models. Among the latter, we consider the Transformer-based models BarcodeBERT, DNABERT, and DNABERT-2, along with SSM-based models including HyenaDNA and Caduceus, selecting versions with comparable parameter counts available on HuggingFace. We adopt the hyperparameter settings reported for these models and conduct a grid search over linear probing hyperparameters, including the learning rate, momentum, and weight decay for the SGD optimizer. Specifically, we test learning rates of $[0.01, 0.1, 0.5]$, momenta of $[0.2, 0.4, 0.6, 0.8]$ and weight decays of $[10^{-8}, 10^{-9}, 10^{-11}]$. Finally, we present the best results for all baselines and compare them with the performance of BarcodeMamba in Table \ref{comparison}.

\paragraph{Results}
The results of our comparison study are presented in Table \ref{comparison}. In fine-tuning (first column) we see that all models perform reasonably well, with HyenaDNA-tiny achieving the highest accuracy by a small margin. However, in the more challenging test of SSL-trained representations (columns 2 \& 3), similar to BarcodeBERT, our linear and 1-NN probing results demonstrate a substantial improvement compared to all other models. In terms of parameters, our BarcodeMamba model exhibits superior performance to BarcodeBERT with less than 7.4\,M parameters (vs.~86.2-89.2\,M). Utilizing the character-level tokenizer and NTP pretraining objective, BarcodeMamba achieves high accuracy in fine-tuning and linear probing tasks. For the 1-NN probing task, our model benefits from a $k$-mer tokenizer with $k=6$. As we scale up BarcodeMamba to 56.7\,M parameters, it reaches the highest accuracy in linear probing as well as 1-NN probing, indicating great potential for practical biodiversity analysis.

\begin{table}[h!]
\caption{Two groups of baselines: off-the-shelf foundation models pretrained on human genome datasets vs.~BarcodeBERT and our model BarcodeMamba, which are specifically pretrained on DNA barcode-based datasets. We sort these models by their number of parameters in descending order within the respective families to facilitate comparison. The numbers in parentheses are the optimal $k$-mer values that yielded the best results, where $k$=1 denotes the use of a character-level tokenizer. The parameter counts are presented as ranges due to the variability in vocabulary sizes associated with different values of $k$. $\uparrow$/$\downarrow$ denotes metrics where higher/lower values are better.}
\centering
\resizebox{\textwidth}{!}{%
\begin{tabular}{@{}lrrrr@{}}
\toprule
 & \multicolumn{2}{l}{\makecell{Species-level acc (\%) $\uparrow$\\ of seen species}} &  \multicolumn{1}{l}{\makecell{Genus-level acc (\%) $\uparrow$\\ of unseen species}} & \multicolumn{1}{l}{} \\ \cmidrule(l){2-4}
Model & Fine-tuned & Linear probe & 1-NN probe & Params $\downarrow$\\ \midrule
DNABERT-2 & 98.3 & 87.2 & 40.9 & 118.9\,M \\
DNABERT & ($k$=6) 97.4 & ($k$=4) 47.1 & ($k$=6) 48.5 & 88.1-91.1\,M \\
Caduceus-PS-131k & 97.6 & 5.1 & 21.1 & 14.0\,M \\
Caduceus-PH-131k & 96.7 & 2.7 & 19.3 & 14.0\,M \\
Caduceus-PS-1k & 98.8 & 16.8 & 31.4 & 3.5\,M \\
Caduceus-PH-1k & 98.8 & 6.2 & 23.1 & 3.5\,M \\
HyenaDNA-small & 98.5 & 75.2 & 46.1 & 3.3\,M \\
HyenaDNA-tiny & \textbf{99.1} & 93.5 & 47.0 & 1.6\,M \\
CNN encoder & 98.2 & 51.8 & 47.0 & 1.8\,M \\ \midrule
BarcodeBERT & ($k$=6) 98.1 & ($k$=4) 93.0 & ($k$=5) 58.4 & 86.2-89.2\,M \\
BarcodeMamba-2-large (ours) & ($k$=6) 97.7 & ($k$=1) \textbf{99.2} & ($k$=6) \textbf{70.2} & 50.4-56.7\,M \\
BarcodeMamba-2-mini (ours) & ($k$=1) 97.7 & ($k$=1) \textbf{99.2} & ($k$=6) 63.2 & 4.3-7.4\,M \\ \bottomrule
\end{tabular}%
}\label{comparison}
\end{table}

\subsubsection{Ablation Study}
\paragraph{Implementation Details}
We evaluated two tokenizers during training and inference: character-level and $k$-mer-based. For $k$-mer length, we adhere to the approach of BarcodeBERT and set $k=4,5,6$. Two pretext tasks for pretraining are explored: NTP and MLM. We use the AdamW optimizer with a learning rate of $6 \times 10^{-4}$, a weight decay of 0.1, and betas set to 0.9 and 0.999. A cosine learning rate scheduler is applied, which includes a small learning rate that linearly warms up over the first 1\% of the training duration before decaying to 10\% of the initial learning rate.

In terms of BarcodeMamba's architecture, we set the model dimension to $d=256$, number of layers to $n=2$, and head dimension to $p=64$.
\paragraph{Results}
As demonstrated in Tables \ref{ablationntp} and \ref{ablationmlm}, for both pretraining tasks, Mamba-2 performs better as the mixing layer in most scenarios.
When using NTP, as detailed in Table \ref{ablationntp}, utilizing a character-level tokenizer enhances the fine-tuning and linear probing outcomes of BarcodeMamba. This suggests that character-level tokenization contributes to improved representation learning for the task at hand. However, for 1-NN probing, the $k$-mer tokenization enables BarcodeMamba to achieve significantly better results than character-level tokenization. Furthermore, as the length of the $k$-mer increases, the accuracy of probing on unseen datasets improves. This indicates that $k$-mer-based tokenization captures shared motifs and sub-sequences across seen and unseen species' barcodes more effectively with larger window sizes. During testing with pretrained models on an unseen dataset, BarcodeMamba generally shows higher perplexity with $k$-mer tokenization compared to character-level tokenization. This can be attributed to the fact that there are only 5 characters in the vocabulary for the character-level tokenizer, compared to $4^{k}+\texttt{n\_special\_tokens}$ vocabulary size for the $k$-mer tokenizer.

The advantage of using a character-level tokenizer with MLM (Table \ref{ablationmlm}) is not as substantial as with NTP (Table \ref{ablationntp}). Although BarcodeMamba achieves a lower perplexity on the unseen dataset, the results on linear probing and 1-NN probing are reduced by approximately 2--3 points. Despite this, BarcodeMamba remains performant for the fine-tuning task with Mamba-2 as the mixing layer, demonstrating similar performance to NTP.

\begin{table}[h!]
\caption{Classification Accuracy and Pretraining Perplexity of BarcodeMamba in Different Settings with NTP: We present results using a character-level and $k$-mer tokenizer under various settings, focusing on the impact of different $k$-mer lengths (i.e., $k=4,5,6$). Perplexity scores are comparable within a row but not across rows because of the changing vocabulary size. Therefore those are not bolded. $\uparrow$/$\downarrow$ denotes metrics where higher/lower values are better.}
\centering
\resizebox{\textwidth}{!}{%
\begin{tabular}{@{}lccccccccc@{}}
\toprule
\multicolumn{1}{c}{} &  & \multicolumn{4}{c}{\makecell{Species-level acc (\%) $\uparrow$\\ of seen species}} & \multicolumn{2}{c}{\makecell{Genus-level acc (\%) $\uparrow$\\ of unseen species}} & \multicolumn{2}{c}{\makecell{Representation \\ of unseen barcodes}} \\ \cmidrule(l){3-6}\cmidrule(l){7-8}
\cmidrule(l){9-10}
\multicolumn{1}{c}{} &  & \multicolumn{2}{c}{Fine-tuned} & \multicolumn{2}{c}{Linear probe} & \multicolumn{2}{c}{1-NN probe} & \multicolumn{2}{c}{Perplexity $\downarrow$} \\ \cmidrule(l){3-4} \cmidrule(l){5-6} \cmidrule(l){7-8} \cmidrule(l){9-10}
Tokenizer & $ k $ & Mamba & Mamba-2 & Mamba & Mamba-2 & Mamba & Mamba-2 & Mamba & Mamba-2 \\ \midrule
Char & - & \textbf{98.7} & \textbf{98.1} & \textbf{97.0} & \textbf{95.9} & 41.2 & 33.0 & 1.41 & 1.37 \\
$ k $-mer & 4 & 95.0 & 97.4 & 92.9 & 94.0 & 43.5 & 55.3 & 3.19 & 3.09 \\
$ k $-mer & 5 & 94.2 & 95.6 & 91.5 & 92.6 & \textbf{48.5} & 57.7 & 4.16 & 4.04 \\
$ k $-mer & 6 & 95.9 & 96.5 & 91.8 & 91.9 & 47.7 & \textbf{58.7} & 5.51 & 5.31 \\ \bottomrule
\end{tabular}%
}\label{ablationntp}
\end{table}

\begin{table}[h!]
\caption{Classification Accuracy and Pretraining Perplexity of BarcodeMamba in Different Settings with MLM: We present results using a character-level and $k$-mer tokenizer under various settings, focusing on the impact of different $k$-mer lengths (i.e., $k=4,5,6$). Perplexity scores are comparable within a row but not across rows because of the changing vocabulary size. Therefore those are not bolded. $\uparrow$/$\downarrow$ denotes metrics where higher/lower values are better.}
\centering
\resizebox{\textwidth}{!}{%
\begin{tabular}{@{}lccccccccc@{}}
\toprule
\multicolumn{1}{c}{} &  & \multicolumn{4}{c}{\makecell{Species-level acc (\%) $\uparrow$\\ of seen species}} & \multicolumn{2}{c}{\makecell{Genus-level acc (\%) $\uparrow$\\ of unseen species}} & \multicolumn{2}{c}{\makecell{Representation \\ of unseen barcodes}} \\ \cmidrule(l){3-6}\cmidrule(l){7-8}
\cmidrule(l){9-10}
\multicolumn{1}{c}{} &  & \multicolumn{2}{c}{Fine-tuned} & \multicolumn{2}{c}{Linear probe} & \multicolumn{2}{c}{1-NN probe} & \multicolumn{2}{c}{Perplexity $\downarrow$} \\ \cmidrule(l){3-4} \cmidrule(l){5-6} \cmidrule(l){7-8} \cmidrule(l){9-10}
Tokenizer & $ k $ & Mamba & Mamba-2 & Mamba & Mamba-2 & Mamba & Mamba-2 & Mamba & Mamba-2 \\ \midrule
Char & - & 88.4 & \textbf{98.2} & 91.8 & 91.5 & 32.1 & 38.7 & 1.23 & 1.22 \\
$ k $-mer & 4 & \textbf{97.3} & 96.6 & \textbf{94.0} & \textbf{94.3} & 47.4 & 50.4 & 1.89 & 1.86 \\
$ k $-mer & 5 & 97.1 & 97.5 & 92.9 & 93.1 & 52.2 & \textbf{51.9} & 2.20 & 2.17 \\
$ k $-mer & 6 & 96.7 & 95.4 & 92.7 & 92.7 & \textbf{54.5} & 51.0 & 2.46 & 2.45 \\ \bottomrule
\end{tabular}%
}\label{ablationmlm}
\end{table}
\vspace{-0.2cm}
\subsubsection{Scaling up BarcodeMamba}
\paragraph{Implementation Details}
Based on the results of our ablation study, we scaled up BarcodeMamba with the NTP pretraining objective in both character-level and $k$-mer tokenizer settings ($k=6$), as these configurations showed the most promise in fine-tuning accuracy, linear probing, and 1-NN probing accuracy. Details on the number of layers, model dimensions, and batch sizes are provided in Table~\ref{scalingconfig}. BarcodeMamba uses more memory when using a character-level tokenizer due to the increased sequence length required for learning barcode representations at single nucleotide resolution. We implemented an early stopping approach with a maximum epoch limit of 25, pretraining all models for 8.03--42.41 hours and fine-tuning them in less than 1 hour.

\begin{table}[h!]
\caption{Model configurations for scaling up BarcodeMamba. The panel on the left shows configurations that were systematically chosen. The panel on the right displays the optimal hyperparameters for probing accuracy.}\label{scalingconfig}
\begin{minipage}{0.5\linewidth}\scalebox{0.9}{
\begin{tabular}{@{}rrrrrr@{}}
	\toprule
	 \multicolumn{2}{c}{Model size} & \multicolumn{2}{c}{Batch size} & \multicolumn{2}{c}{Params} \\ \cmidrule
	(l){1-2}\cmidrule(l){3-4}\cmidrule(l){5-6}
	Layers $n$ & Dim $d$ & Char & $k$-mer & Char & $k$-mer \\ \midrule
	 2 & 256 & 256 & 256 & 1.9\,M & 4.0\,M \\
	 8 & 256 & 256 & 256 & 7.7\,M & 9.8\,M \\
	 8 & 512 & 128 & 256 & 30.1\,M & 34.2\,M \\
	 10 & 768 & 64 & 256 & 87.2\,M & 90.2\,M \\ \bottomrule
\end{tabular}}
\end{minipage}
\begin{minipage}{0.5\linewidth}\scalebox{0.9}{
\begin{tabular}{@{}rrrrrr@{}}
	\toprule
	 \multicolumn{2}{c}{Model size} & \multicolumn{2}{c}{Batch size} & \multicolumn{2}{c}{Params} \\ \cmidrule
	(l){1-2}\cmidrule(l){3-4}\cmidrule(l){5-6}
	Layers $n$ & Dim $d$ & Char & $k$-mer & Char & $k$-mer \\ \midrule
	 2 & 384 & 16 & 16 & 4.3\,M & 7.4\,M \\
	 4 & 512 & 16 & 16 & 15.0\,M & 19.2\,M \\
	 4 & 768 & 16 & 16 & 33.6\,M & 39.9\,M \\
	 6 & 768 & 16 & 16 & 50.4\,M & 56.7\,M \\ \bottomrule
\end{tabular}}
\end{minipage}
\end{table}
\vspace{-0.4cm}
\paragraph{Results}

\begin{figure}[h!]
\caption{Scaling analysis: Classification accuracy (\%) of BarcodeMamba using a pretrained model as a feature extractor. Metrics are compared between pretraining with a character-level tokenizer and with a $k$-mer tokenizer ($k=6$) (as labeled in the sub-figures). LP represents Linear Probe and 1-NN is short for 1-Nearest Neighbor Probe.} \label{scaling_figs}
\begin{minipage}{0.5\linewidth}
\includegraphics[width=\linewidth]{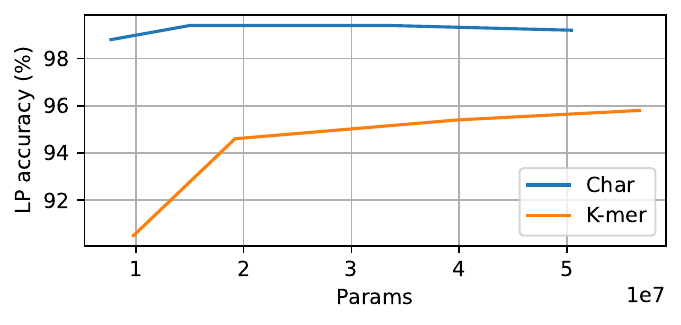}
\end{minipage}
\begin{minipage}{0.5\linewidth}
\includegraphics[width=\linewidth]{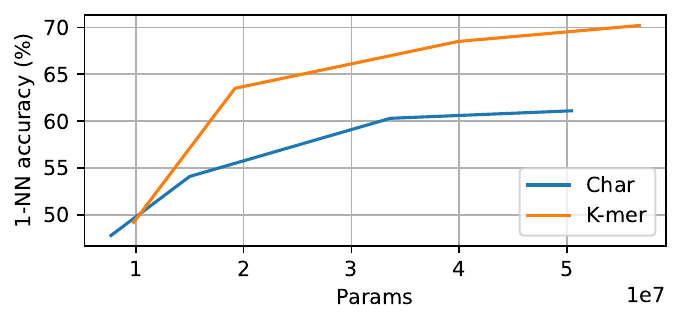}
\end{minipage}
\end{figure}
\vspace{-0.4cm}

\begin{table}[h!]
\caption{The evaluation of the BarcodeMamba performance involves perplexity and classification accuracy, using a pretrained model as a feature extractor. Metrics are compared between pretraining with a character-level tokenizer (left) and with a $k$-mer tokenizer ($k=6$) (right). FT stands for Fine-Tuning, LP represents Linear Probe and 1-NN is short for 1-Nearest Neighbor Probe. $\uparrow$/$\downarrow$ denotes metrics where higher/lower values are better.}
\begin{minipage}{0.5\linewidth}
\scalebox{0.79}{\begin{tabular}{@{}ccccr@{}}
\toprule
Perplexity $\downarrow$ & FT(\%) $\uparrow$ & LP(\%) $\uparrow$ & 1-NN(\%) $\uparrow$ & Params $\downarrow$ \\ \midrule
1.36 & 98.1 & 98.4 & 40.6 & 1.9\,M \\
1.36 & 97.7 & 99.2 & 57.9 & 4.3\,M \\
1.32 & 97.9 & 98.8 & 47.8 & 7.7\,M \\
1.34 & 95.4 & \textbf{99.4} & 54.1 & 15.0\,M \\
1.31 & 97.7 & \textbf{99.4} & 59.3 & 30.1\,M \\
1.32 & 98.3 & \textbf{99.4} & 60.3 & 33.6\,M \\
1.28 & 94.9 & 99.2 & \textbf{61.1} & 50.4\,M \\
\textbf{1.27} & \textbf{98.2} & 99.3 & 58.5 & 87.2\,M \\ \bottomrule
\end{tabular}}
\end{minipage}
\begin{minipage}{0.5\linewidth}
\scalebox{0.79}{\begin{tabular}{@{}ccccr@{}}
\toprule
Perplexity $\downarrow$ & FT(\%) $\uparrow$ & LP(\%) $\uparrow$ & 1-NN(\%) $\uparrow$ & Params $\downarrow$ \\ \midrule
5.34 & 96.2 & 91.9 & 58.5 & 4.0\,M \\
\textbf{5.07} & 96.9 & 93.6 & 63.2 & 7.4\,M \\
5.10 & 91.5 & 90.5 & 49.2 & 9.8\,M \\
5.20 & 95.7 & 94.6 & 63.5 & 19.2\,M \\
5.32 & 93.6 & 94.0 & 60.4 & 34.2\,M \\
5.34 & 95.9 & 95.4 & 68.5 & 39.9\,M \\
5.43 & \textbf{97.7} & \textbf{95.8} & \textbf{70.2} & 56.7\,M \\
5.55 & 94.7 & 94.7 & 60.5 & 90.2\,M \\ \bottomrule
\end{tabular}}
\label{scaling}
\end{minipage}
\end{table}

The visualization depicted in Figure \ref{scaling_figs} demonstrates that under optimal model dimensions and number of layers, both linear and 1-NN probing accuracy increase as the parameter count of BarcodeMamba increases. Furthermore, Table \ref{scaling} provides a comprehensive set of scaling results for all metrics, including Perplexity and Fine-tuning, showing how the effectiveness of NTP and classification performance change as models grow in number of parameters.
The performance of BarcodeMamba with a character-level tokenizer is shown in Table \ref{scaling} (left), where perplexity, fine-tuning, seen species-level and unseen genus-level probing accuracy improve as BarcodeMamba scales up. Specifically, the linear probing accuracy reaches a peak of 99.4\%, while the 1-NN probing accuracy achieves its highest value of 61.1\% at 50.4\,M parameters.
As shown in Table \ref{scaling} (right), scaling up the BarcodeMamba model with the $k$-mer tokenizer ($k=6$) improves its classification performance in linear and 1-NN probing. Overall, BarcodeMamba shows potential for discovering new species in biodiversity research, as it scales effectively in the zero-shot 1-NN probing task.

As we scaled up BarcodeMamba, we observed a slight overfit in models that use the $k$-mer tokenizer based on perplexity. Scaling from 4\,M to 90\,M parameters, models resulted in train perplexities of 1--2. However, the test perplexity remained above 5.07 for all models. While these results demonstrate BarcodeMamba's effectiveness in DNA barcode analysis, they also suggest room for further improvements through increased training data and enhanced data augmentation strategies. Therefore, our future work will explore extending the use of BarcodeMamba beyond the Canadian invertebrate dataset and evaluating its performance on BIOSCAN-5M \cite{bioscan5m}, a recently-released extensive biodiversity dataset with 5 million insect specimens.

\section{Conclusions}
We demonstrate that Mamba-2-based models pretrained with next token prediction on DNA barcode data achieve high performance in arthropod species identification while maintaining computational efficiency. Through comprehensive experiments comparing architectures, ablating components, and analyzing scaling behaviour, we explored how pretraining objectives and tokenization methods affect SSM-based foundation models. Our results show that BarcodeMamba achieves strong performance in taxonomic classification of both seen and unseen species, demonstrating its potential for biodiversity science. Future work will focus on scaling BarcodeMamba to the larger and more taxonomically diverse BIOSCAN-5M dataset to further improve species identification performance. We will also explore architectural modifications, including bi-directional variants, to enhance the model's capabilities for biodiversity analysis.

\begin{ack}
Iuliia Zarubiieva, Scott C.~Lowe, and Pablo Millan Arias read drafts of the manuscript and provided valuable feedback. BIOSCAN is supported in part by funding from the Government of Canada's New Frontiers in Research Fund (NFRF). Resources used in preparing this research were provided, in part, by the Province of Ontario, the Government of Canada through the Canadian Institute for Advanced Research (CIFAR), and companies sponsoring the Vector Institute \url{http://www.vectorinstitute.ai/\#partners}. GWT acknowledges support from the Natural Sciences and Engineering Research Council (NSERC), the Canada Research Chairs program, and the Canada CIFAR AI Chairs program.
\end{ack}

\small
\bibliographystyle{plainnat}
\bibliography{references}

\end{document}